\documentclass[10pt, a4paper]{article}
\usepackage{lrec2022} 
\usepackage{multibib}
\newcites{languageresource}{Language Resources}
\usepackage{graphicx}
\usepackage{tabularx}
\usepackage{soul}
\usepackage[dvipsnames]{xcolor}

\usepackage{titlesec}
\titleformat{\section}{\normalfont\large\bfseries\center}{\thesection.}{1em}{}
\titleformat{\subsection}{\normalfont\SmallTitleFont\bfseries\raggedright}{\thesubsection.}{1em}{}
\titleformat{\subsubsection}{\normalfont\normalsize\bfseries\raggedright}{\thesubsubsection.}{1em}{}
\renewcommand\thesection{\arabic{section}}
\renewcommand\thesubsection{\thesection.\arabic{subsection}}
\renewcommand\thesubsubsection{\thesubsection.\arabic{subsubsection}}

\usepackage{epstopdf}
\usepackage[utf8]{inputenc}

\usepackage{hyperref}
\usepackage{xstring}

\usepackage{color}

\usepackage{booktabs}
\usepackage{url}
\usepackage{textcomp}
\usepackage{graphicx}
\usepackage{amsmath}
\usepackage{multirow} 
\usepackage{enumitem}
\usepackage{xcolor}
\usepackage{authblk}

\usepackage{arydshln}
\newcommand{\bart}{\textsc{Bart}}
\newcommand{\tldr}{\textsc{Tldr}}

\newcommand{\rgl}{\textsc{Rouge-L}}

\newcommand{\rgf}{\textsc{Rouge (F1)}}
\newcommand{\mentsum}{\textsc{MentSum}}

\title{ \textbf{MentSum: A Resource for Exploring \\Summarization of Mental Health Online Posts}}

\name{Sajad Sotudeh$^{\text{1,2}}$*\thanks{\hspace{-0.7em} *Equal contribution}, Nazli Goharian$^{\text{1,2}}$*\footnotemark[1], Zachary Young$^\text{2}$} 

\address{$^\text{1}$Information Retrieval Lab, Georgetown University \\
         $^\text{2}$Department of Computer Science, Georgetown University \\
        {\fontfamily{cmtt}\selectfont{\{sajad, nazli\}@ir.cs.georgetown.edu, zjy2@georgetown.edu}}\\
        }

\abstract{
Mental health remains a significant challenge of public health worldwide. With increasing popularity of online platforms, many use the platforms to share their mental health conditions, express their feelings, and seek help from the community and counselors. Some of these platforms, such as Reachout, are dedicated forums where the users register to seek help. Others such as Reddit provide subreddits where the users publicly but anonymously post their mental health distress. Although posts are of varying length, it is beneficial to provide a short, but informative summary for fast processing by the counselors. To facilitate research in summarization of mental health online posts, we introduce \underline{\textbf{Ment}}al Health \underline{\textbf{Sum}}marization dataset,  \mentsum{}, containing over 24k carefully selected user posts from Reddit, along with their short user-written summary (called \tldr) in English from 43 mental health subreddits. This domain-specific dataset could be of interest not only for generating short summaries on Reddit, but also for generating summaries of posts on the dedicated mental health forums such as Reachout. We further evaluate both extractive and abstractive state-of-the-art summarization baselines in terms of \textsc{Rouge} scores, and finally conduct an in-depth human evaluation study of both user-written and system-generated summaries, highlighting challenges in this research. 
\\ \newline  }

\begin{document}

\maketitleabstract


\section{Introduction}

\begin{figure}[t]
    \centering
    \includegraphics[scale=0.4]{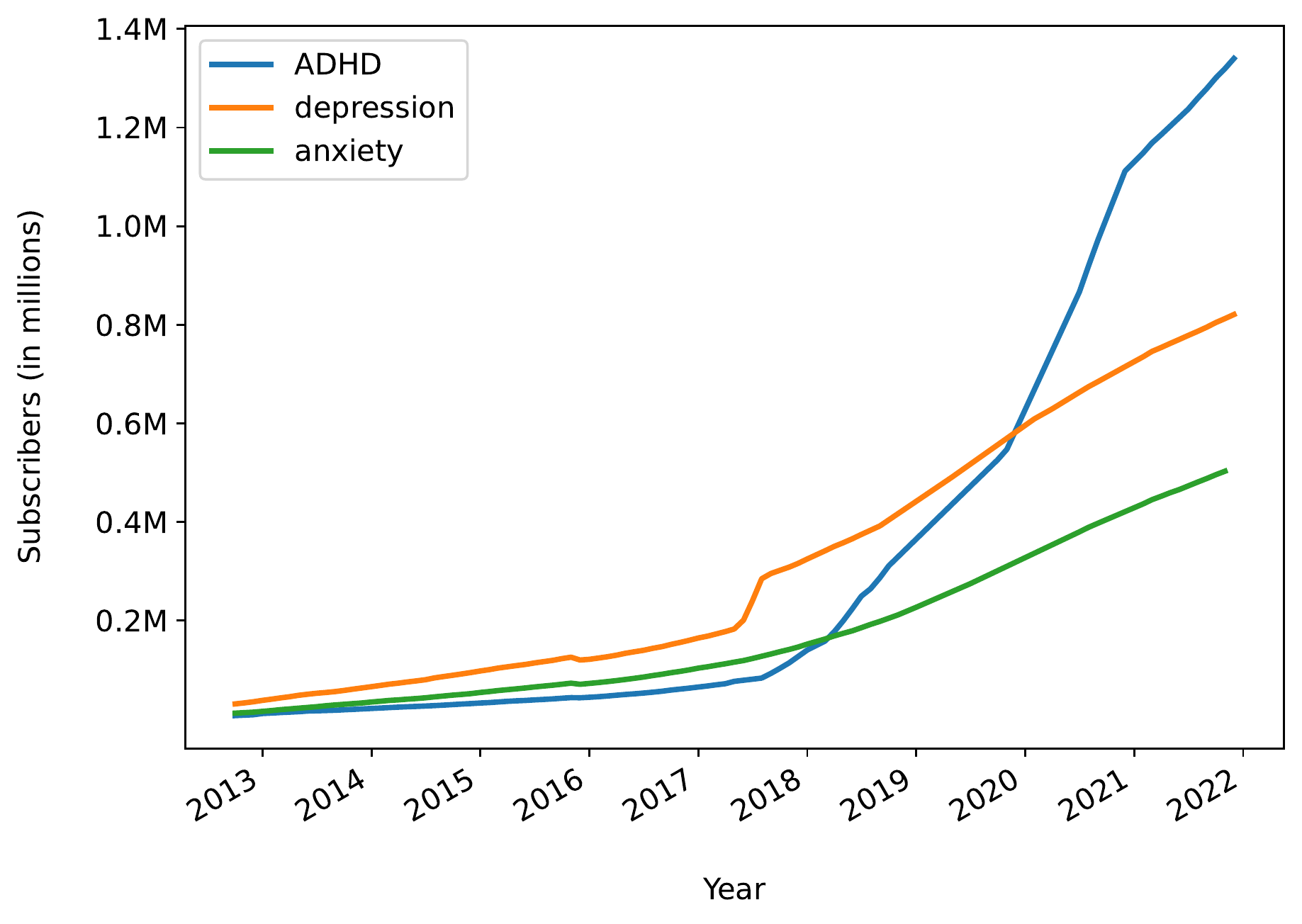}
    \caption{Growing popularity of mental health related forums in Reddit. }
    \label{fig:popularity}
\end{figure}

Mental health has been a global public health challenge for many years and even more so since the COVID-19 pandemic\textcolor{black}{~\cite{Holmes2020MultidisciplinaryRP,Pfefferbaum2020MentalHA,Otu2020MentalHA}.} 
Social media has served as a viable platform for many to share their frustrations, emotions, depressions, and also their already diagnosed mental disorders. Figure \ref{fig:popularity} depicts the growing popularity (\textcolor{black}{measured by the number of subscribers}) of discussion forums dedicated to three mental disorders in Reddit social discussion website \textcolor{black}{over the years}.~\footnote{Statistics  from \url{https://subredditstats.com/}}.



Online social platforms such as \textit{Reddit}~\footnote{\url{https://www.reddit.com/}} and \textit{Reachout}~\footnote{\url{https://au.reachout.com/}} have become increasingly popular over the recent years due to the vital networking facets that they offer to the community users. These platforms provide users with an opportunity to share different types of user-curated and user-generated content, ranging from daily updates/statuses to sharing personal anecdotes and mental conditions. Users can also interact with other users, carry on conversations through which they can express their feelings and views regarding a specific topic. Platforms such as \textit{Reachout} are not public, \textcolor{black}{requiring} users to register; users' content are not visible to anyone but to the permitted users and counselors. On the other hand, in the public platforms such as \textit{Reddit}, users can openly exchange information with each other through community-based \textit{subreddits}, each of which specified with a certain theme or condition, such as suicide watch, mental health, alcoholism, attention-deficit/hyperactivity disorder (ADHD), depression, anxiety, etc. Each post in any of these subreddits, however, {may report more than one past} or present condition and what the user is distressed about.  

The user-generated content on many of such platforms might be of varying length. Longer posts may address \textcolor{black}{{multitude of issues of concern} or simply be a lengthy elaboration of the user on the situation.} The longer a post is, the more time it requires a counselor for reading the post which leads to fatigue and/or delay in a timely response.  Our hypothesis is that a short yet informative summary of each user's post provides the counselors with the important information of the post in a glimpse before reading the details. Hence, in this research we create a dataset resource for the research community to be utilized in the short text (known as \tldr) summarization of mental health related social media posts. 
 

A great deal of research studies in social media mental health domain have focused on developing classification models and their needed datasets to either triage the severity of the potential harm or to identify the type of mental disorders; among these efforts are~\cite{Choudhury2013PredictingDV,Coppersmith2014QuantifyingMH,Yates2017DepressionAS,Coppersmith2018NaturalLP,Cohan2018SMHDAL,Garg2021DetectingRL}.
Our goal is not to \textcolor{black}{undermine} classification of disorders but potentially serve as yet another additional form of \textcolor{black}{guidance} to the readers/counselors of the posts through short summaries of the posts. To this end, we had to create a domain-specific dataset to contain social media mental health related posts, along with their short user-written summary as the gold standard. This dataset, \mentsum, contains over 24k posts with pairwise user-written summaries. We hope that \mentsum{} would expedite future work in the social media mental health text summarization task. In short, our contributions are:

\begin{itemize}
    \item Creating Mental Health Summarization dataset, called \mentsum{} that includes \textcolor{black}{over} 24k user posts from the online mental health discussion forums along with their user-written short summaries. 
    \item Providing the results of existing strong baselines in summarization, covering both extractive and abstractive approaches.
    \item Carrying out a human evaluation and error analysis in terms of fluency, informativeness, and conciseness of the system generated summaries, on its own, and in comparison with the user-written summaries. 
    
    We believe that insights from our human evaluation study can be further used by future work to investigate more sophisticated models for moving the field forward. 
\end{itemize}

\section{Related work} 

Online social platforms provide a considerable wealth of textual data, attracting attention of those who study the users' mental conditions in social environments. 
Early works in mental health research have put their focus on understanding and identifying mental health conditions in social media platforms such as Reddit and Twitter~\cite{Choudhury2013PredictingDV,Resnik2013UsingTM,Coppersmith2014MeasuringPT,Mowery2017UnderstandingDS}; particularly, predicting mental state of the users in online media ~\cite{Hao2013PredictingMH,Wang2013AnIM,Mowery2017FeatureST}, exploring the types of mental health condition~\cite{Wilson2014FindingIA}, detecting the severity level of mental disorders~\cite{ODea2015DetectingSO,Chancellor2016QuantifyingAP}, studying mental health discourse~\cite{Choudhury2014MentalHD}, studying language of users and identifying those with a high risk of mental illness~\cite{Milne2016CLPsych2S,Cohan2017TriagingCS}, and analyzing the impact of conversation between a target user and participants~\cite{soldaini-etal-2018-helping}. Some research focused on creating large-scale mental health datasets such as \textsc{rsdd}~\cite{Yates2017DepressionAS}, \textsc{smhd}~\cite{Cohan2018SMHDAL} to detect potential of mental health conditions through the general language of users, and \textsc{rsdd-time}~\cite{MacAvaney2018RSDDTimeTA} to study the temporal information of diagnoses. Unlike existing work whose main focus has been on classification tasks, we define a text summarization task over users' mental health content in online social media platforms. 

While summarization of clinical reports has already attracted the attention of researchers~\cite{Mishra2014TextSI,Goldstein2016AnAK,MacAvaney2019OntologyAwareCA,Zhang2020OptimizingTF,Sotudeh2020AttendTM}, the summarization of social media mental health posts has not been explored previously, which could be due to lack of large-scale mental health summarization datasets. The closest work but yet different than ours is done by \newcite{Manas2021KnowledgeInfusedAS} that aim to summarize the mental health diagnostic interviews on a small-scale conversational dataset (189 patient interviews) without human-written gold summaries. Hence, to the best of our knowledge, we are the first to propose a relatively large scale mental health text summarization dataset based off social media users' content with user-written short summaries.
\section{\mentsum{} dataset}
In this section, we elaborate on construction of our dataset, provide dataset statistics, and analyze the characteristics of the data. Subsequently, we provide the ethics and privacy of the dataset. 

\subsection{Dataset construction}
\label{sec:dataset_cons}
Our constructed dataset is based off Reddit mental health related posts of the users along with their user-written short summaries (called \tldr) as the ground-truth. Note that the author of the post and \tldr{} is the same; hence, the goldness of this ground-truth \tldr{}, in respect to its fluency, and completeness might be impacted by the emotional state of the post's author. The choice of Reddit for building our dataset is motivated by being a public \textcolor{black}{ and popular platform}, namely its public content and also the availability of the short summary ground-truth for each post. Reddit is a social media platform that supports communities called \textit{subreddits}, each dedicated to a specific topic.   

We used Pushshift~\cite{Baumgartner2020ThePR} which is a social media data repository containing recent and historical dumps of posted content on Reddit, which are made publicly available to the Natural Language Processing (NLP) community for research studies. 
We downloaded the Reddit data dumps covering the period of 2005-2021, and filtered the posts based on a set of pre-defined 43 mental health subreddits
.~\footnote{Subreddits are available at \url{https://ir.cs.georgetown.edu/resources/data/mentsum/}} 
As not all of these users' posts have short summaries (i.e., \tldr), using regular expression, we harvested posts that contain a \tldr{} summary 
as done in ~\cite{volske-etal-2017-tl,sotudeh-etal-2021-tldr9}. The regular expression matches keywords that begin with uncased ``TL'' and end with uncased ``DR'', allowing up to three characters in between.

Social media texts are generally unstructured and noisy in terms of having chunky sentences, and grammatical errors~\cite{Baldwin2013HowNS}. This is due to the fact that users can freely express themselves ~\cite{Liu2016TheGT}. To further preserve high-quality instances, we applied the following filtering using a set of hand-crafted filtering rules listed below. 

\begin{enumerate}
    \item \textbf{Token filtering:} We remove a set of markup characters such as ``{\fontfamily{cmtt}\selectfont \&lt}'', ``{\fontfamily{cmtt}\selectfont \&gt}'', ``{\fontfamily{cmtt}\selectfont amp}''
    , etc. that frequently occur within the harvested instances. URLs are also removed and replaced with ``{\fontfamily{cmtt}\selectfont @http }'' tokens. We further remove all non-ASCII characters that may happen within the social media text; hence, preventing their negative effect in the summarization process. We further replace the user IDs or users' names with ``{\fontfamily{cmtt}\selectfont @user}'' tokens to hinder the possibility of users' identities being disclosed. 
    
    \item \textbf{Instance filtering
    } We define two \textcolor{black}{instance} sampling criteria  which should be met by each instance to be included in the final \mentsum{} dataset. First, we identify the most frequent word bigram of the post's \tldr{}; if it occurs more than 3 times (empirically determined), we exclude the instance from the final dataset, otherwise, we keep the instance. This is based on our observation that \tldr{} summaries with more than 3 identical bigrams contain redundant information, not conveying enough information about the posted content in a short summary. 
    Second, we apply a filtering rule based on the \textit{compression ratio}~\footnote{$\text{Compression ratio} =  \frac{\text{count of words in user's post}}{\text{count of words in \tldr}}$} of  instances. Specifically, we only retain instances whose compression ratio falls in the range of [2-13]~\footnote{[2-13] was decided empirically in our experiments.} (i.e., user's post should be between 2x and 13x longer than the associated \tldr{} summary). 
    This filtering decision is based on the notion that to have short summaries we do not want too small of compression ratio;  to have informative enough summaries we do not want too large of compression ratio.
   
\end{enumerate}

The pipeline that we mentioned above reduced the initial set of 42k instances to the final dataset with 
24,119 English post-\tldr{} user-written summary pairs. %
\begin{table}
    \centering
    
        \begin{tabular}{c}
         
         \resizebox{\columnwidth}{!}{%
    \begin{tabular}{lr}
    \toprule
     Dataset size & 24,119 posts \\
     Training set size & 21,695 posts \\
     Subreddit coverage & 43 subreddits \\ \vspace{-0.3cm}
\\ \hdashline \vspace{-0.3cm}  \\
     Average post length (word/sent.) & 327.5 / 16.9\\
    Min post length (word/sent.) & 50 / 1\\
    Max post length (word/sent.) & 2,979 / 192 \\
    Median post length (word/sent.) & 267.0 / 14.0\\ \vspace{-0.3cm}
\\ \hdashline \vspace{-0.3cm}  \\
     Average \tldr{} length (word/sent.)& 43.5 / 2.6\\
     Min \tldr{} length (word/sent.)& 15 / 1\\
     Max \tldr{} length (word/sent.)& 596 / 36\\
     Median \tldr{} length (word/sent.)& 35.0 / 2.0 \\ \vspace{-0.3cm}
\\ \hdashline \vspace{-0.3cm}  \\
     Average sentence length (post/\tldr)  & 19.4 / 18.3 words \\
     Median sentence length (post/\tldr) & 17.0 / 16.0 words \\ \vspace{-0.3cm}
\\ \hdashline \vspace{-0.3cm}  \\
     Compression ratio (average) & 7.5 \\
     \bottomrule
    \end{tabular}
    }
         \vspace{0.7em}  
         \\
         (a)
         \vspace{1em}
         \\
         
    \begin{tabular}{lr}
    \toprule
     Total vocabulary size & 76,411 words\\
     Occurring 10+ times & 17,486 words\\
     Training vocabulary size & 49,037 words\\
     Validation vocabulary size & 13,609 words\\
     Test vocabulary size & 13,765 words\\
     
     Training/Test vocabulary overlap & 91.4\%\\
     \bottomrule
     
    \end{tabular}
    
        \vspace{0.7em}  
         \\
         (b)
    \end{tabular}

    \caption{(a) Statistics of \mentsum{} dataset. (b) Vocabulary statistics over distinct and uncased vocabulary terms. }
    \label{tab:mentsum_stat}
\end{table}

\subsection{Dataset statistics}

\begin{figure}
    \centering
    \includegraphics[scale=0.2]{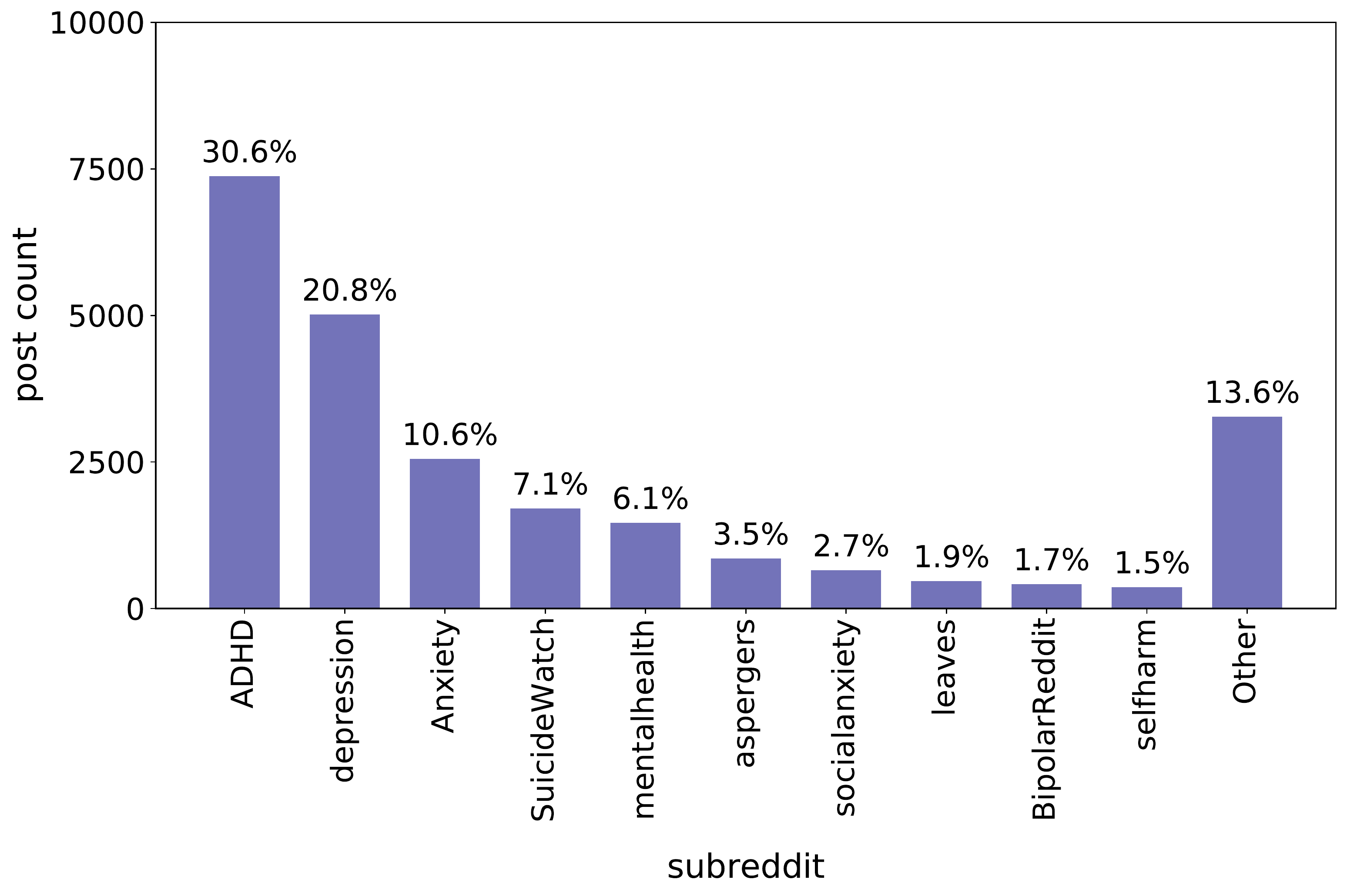}
    \caption{Count (y-axis) and proportion (above each bar) of posts of top 10 frequent subreddits  in \mentsum{} dataset. ``Other'' includes the posts in the remaining 33 mental health subreddits. }
    \label{fig:tldr_precentage}
\end{figure}

The overall dataset statistics are shown in Tables \ref{tab:mentsum_stat} (a). As shown, the compression ratio of this dataset amounts to 7.5 which shows that \tldr{} summaries are the extremely short version of their associated users' post. Table \ref{tab:mentsum_stat} (b) presents the statistics on distinct and uncased vocabularies included in the dataset. As shown, about 23\% of the vocabularies occur more than 10 times within the dataset. Figure \ref{fig:tldr_precentage} depicts the count and percentage of the posts for top 10 subreddits that have the highest count of posts. As indicated, {\fontfamily{cmtt}\selectfont ADHD} has the highest count of posts in \mentsum{}, followed by {\fontfamily{cmtt}\selectfont depression}, {\fontfamily{cmtt}\selectfont anxiety}, {\fontfamily{cmtt}\selectfont SuicideWatch}, and the rest\footnote{While our dataset covers a range of 43 mental health subreddits, we only show top 10 of them in terms of post frequency due to space constraints.}. 

In order to make different sets for training and evaluating summarization models, we randomly divide the data into 21,695 training (90\%), 1,209 validation (5\%) and 1,215 test (5\%) instances.

\begin{figure}
    \centering
    \includegraphics[scale=0.15]{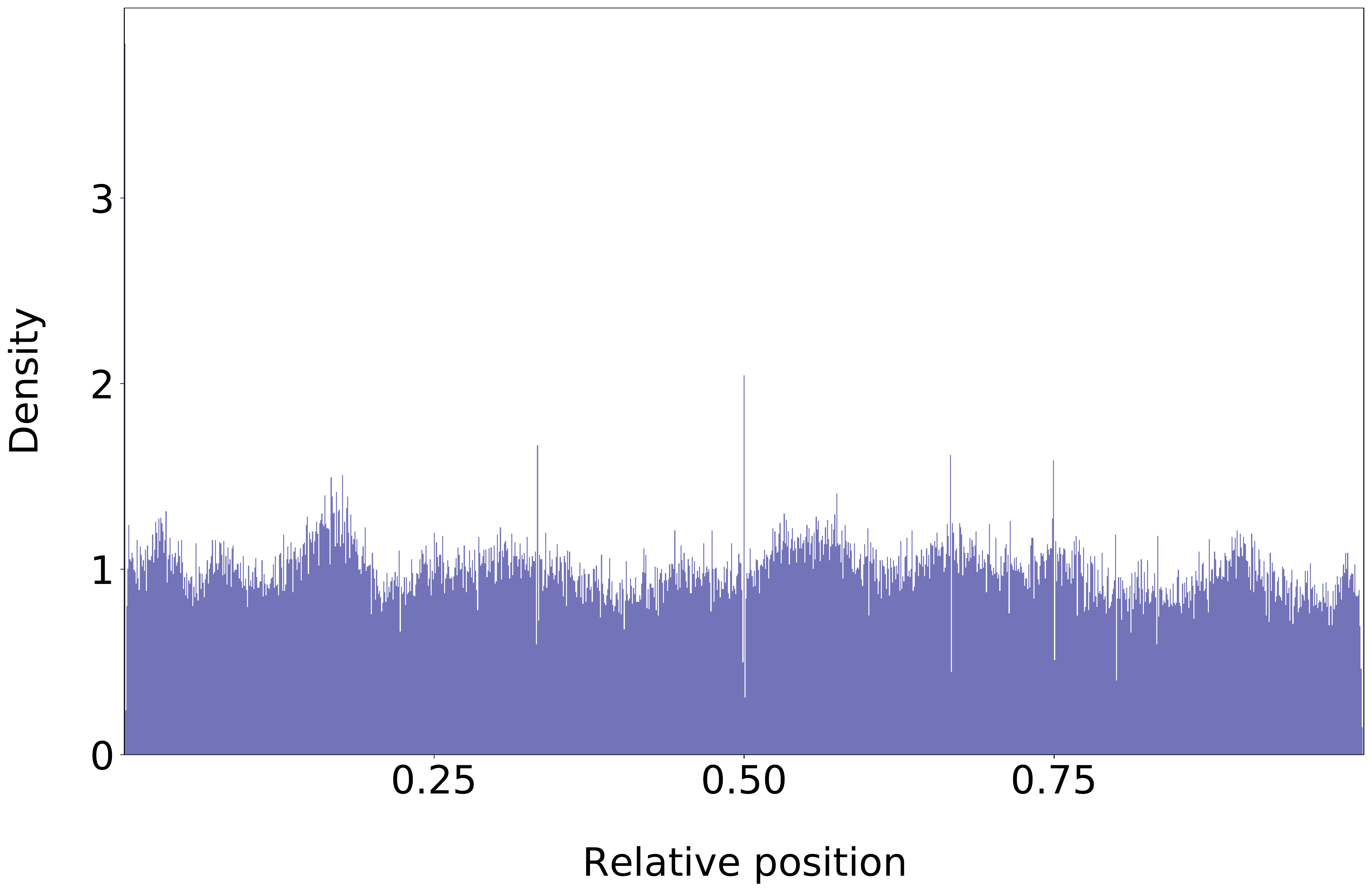}
    \caption{The relative position of word bigrams of the user-written \tldr{} summary across users' post in \mentsum{} dataset.  }
    \label{fig:lead}
\end{figure}

\begin{figure}[t]
    \centering
    \includegraphics[scale=0.35]{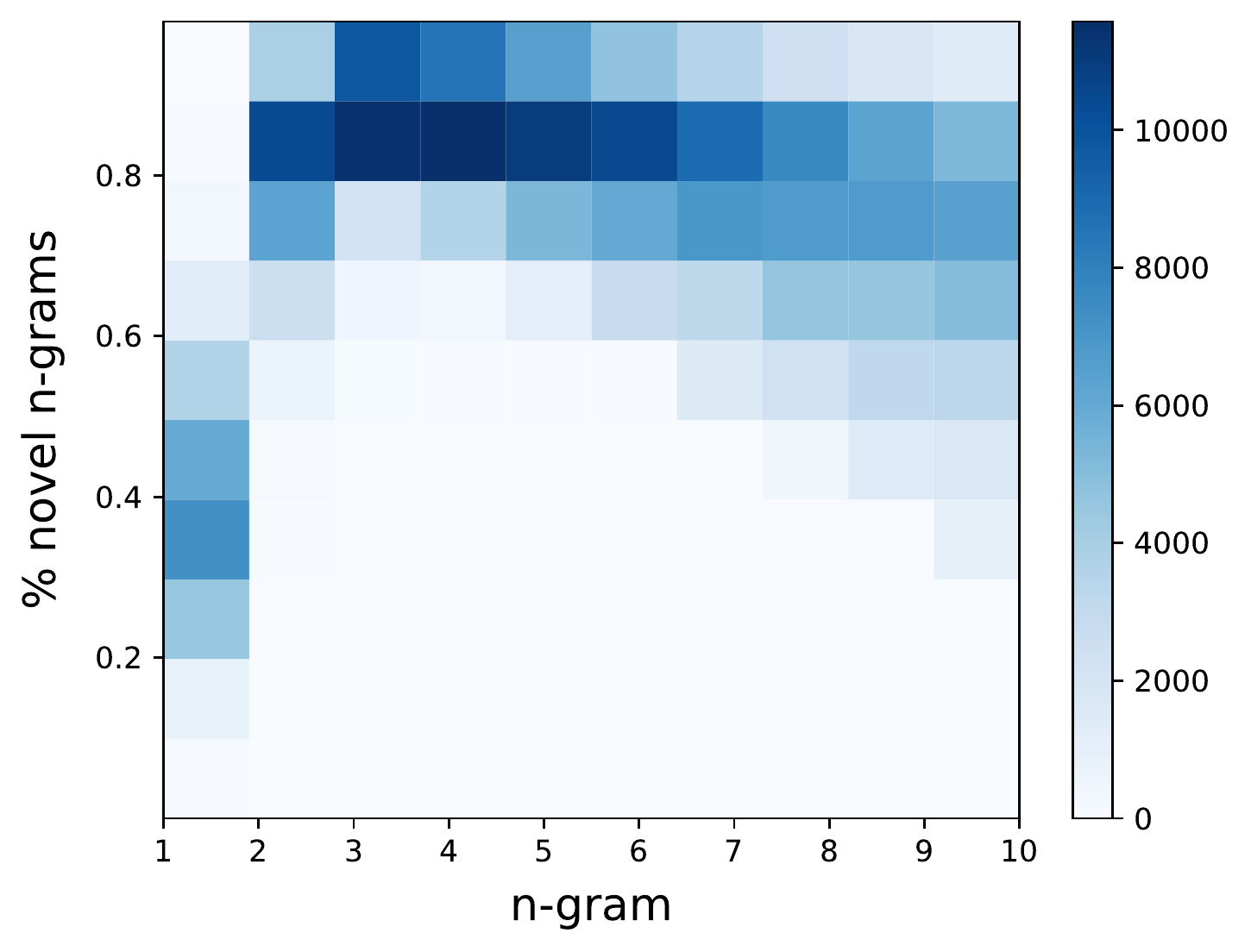}
    \caption{Percentage of novel n-grams (y-axis) across increasing word n-grams (x-axis) in user-written \tldr s. The heat extent shows the frequency of word n-grams for a certain percentage bin.}
    \label{fig:novel}
\end{figure}

\subsection{Dataset analysis}
\label{sec:dataset_analysis}
In this section, we provide an analysis of \mentsum{} dataset in terms of lead bias, abstractive, and extractive characteristics of data.

\textbf{Lead bias. } Lead bias~\cite{Hong2014ImprovingTE} is a common phenomenon in News summarization datasets,
 where the early parts of the source document contain the most salient information. However, this characteristic does not hold in the social media posts as the salient information are scattered in the entire user's post~\cite{Kim2019AbstractiveSO}; hence, imposing a challenge for the summarization task. 
 Figure \ref{fig:lead} demonstrates the relative position of word bigrams of the \tldr{} summary within users' posts in \mentsum{} dataset. As observed, the \tldr{} summary's bigrams are uniformly distributed along the users' post text, exhibiting a weak lead bias. 

\textbf{Abstractiveness.} We further measure the abstractiveness of the purposed \mentsum{} dataset to verify its applicability for abstractive text summarization models. Figure \ref{fig:novel} shows the percentage of novel words for different n-grams. 
As observed, the heat is mainly populated in the upper bins (i.e., higher probabilities), particularly for $\{n | n=2, 3, 4, 5, 6\}$, which shows strong abstractive characteristics of the \mentsum{} dataset, making it suitable for abstractive summarization task.  

\begin{figure}
    \centering
    \includegraphics[scale=0.35]{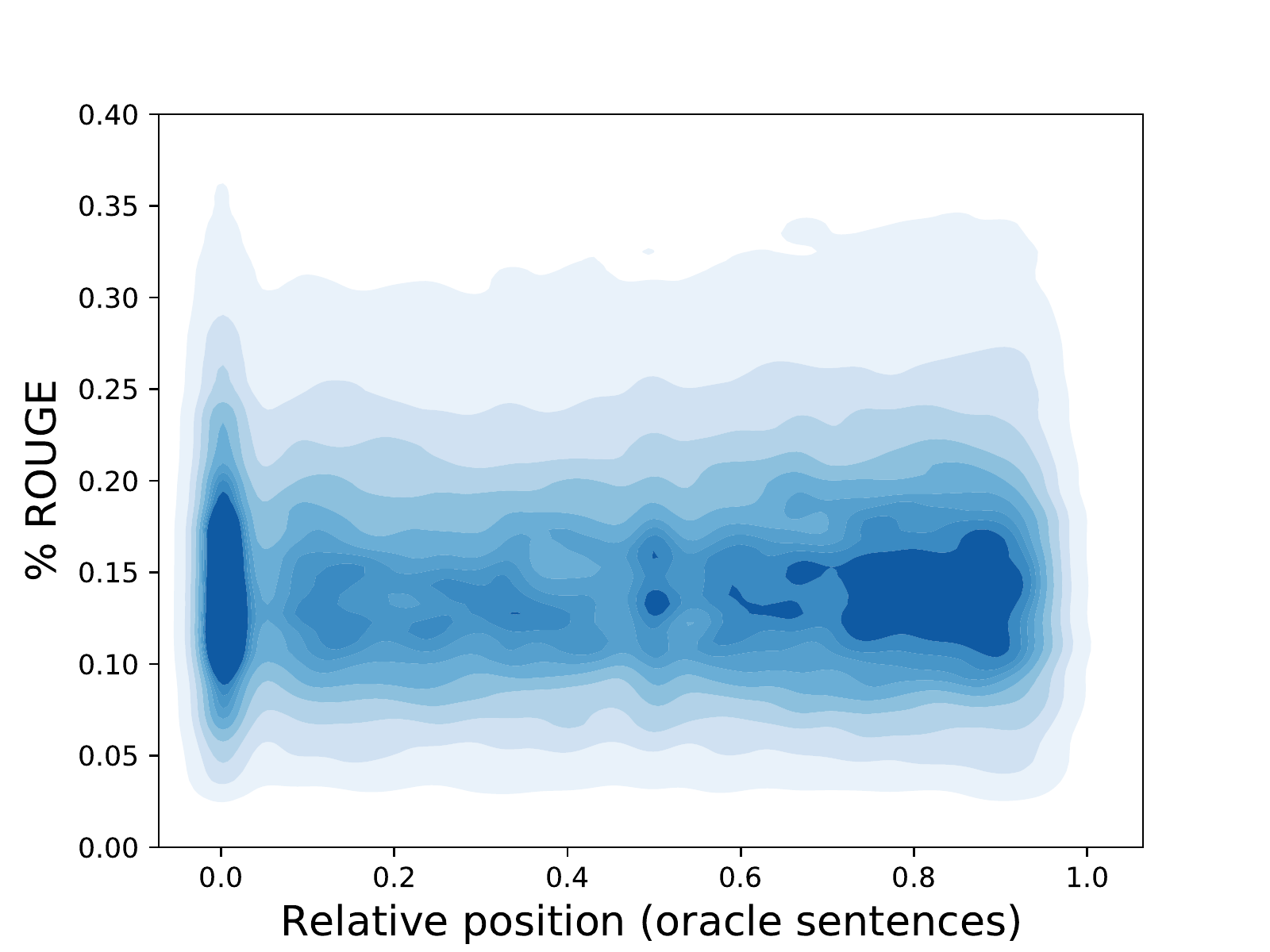}
    \caption{Kernel density estimation diagram of oracle sentences' relative position and the \textsc{Rouge} score of oracle sentences in respect to user-written \tldr s. }
    \label{fig:ext}
\end{figure}

\textbf{Extractiveness.} Figure \ref{fig:ext} shows the density estimation diagram of oracle sentences' relative position in the users' post in regard to the \textsc{Rouge} score~\footnote{We have taken average of \textsc{Rouge-1}, \textsc{Rouge-2}, and \textsc{Rouge-L} scores w.r.t the \tldr{} summary.} of the oracle sentences. Oracle sentences are up to 3 summary-worthy sentences which are labelled using a greedy labeling approach proposed in ~\newcite{Liu2019TextSW}. As indicated, the oracle sentences appear across various positions of the users' post. Considering the diagram and the range of oracle sentences' scores, \mentsum{} dataset still expresses extractive characteristics in addition to its high abstractiveness.  

\subsection{Ethics and privacy}
 Although we use publicly available Reddit data in our research to construct the \mentsum{} dataset, mental health is a sensitive topic and special care should be taken when such data is used in social media research~\cite{Thomas2002ACO,Moreno2013EthicsOS,Suster2017ASR,Benton2017EthicalRP,Nicholas2020EthicsAP}. Hence, we made no attempt to identify and contact the users, or discover user relations with other social media accounts. In preprocessing step, we have de-identified the usernames with \texttt{@user} tokens to prevent the user identities from being known. The \mentsum{} dataset can be accessed through a Data Usage
Agreement (DUA)~\footnote{\url{https://ir.cs.georgetown.edu/resources/}}. The DUA particularly ensures that no attempts should be made to distribute portions of dataset (which could result in revealing users' identity), identify users, and contact users.

\section{Experimental setup}
To evaluate the quality of \mentsum{} dataset for the summarization task of mental health related posts, and to provide strong baselines for further research, we explored several baselines. In this section, we present the baselines and the implementation details. 

\subsection{Baselines}
We explored various extractive and abstractive baselines which are listed below. 
\begin{itemize}[leftmargin=*,label={-}]
\item \textsc{\textbf{Lead-2}}: A simple extractive baseline that selects the first two leading sentences as the summary. 

\item \textsc{\textbf{LSA}}~\cite{Steinberger2004LSA}: A non-neural extractive vector-based model that adopts the mathematical concept of Singular Value Decomposition (SVD) to find hidden semantic structures of words and sentences.

\item \textsc{\textbf{LexRank}}~\cite{Erkan2004LexRankGL}: An unsupervised extractive model that makes use of graph centrality network to find important sentences and concatenate them to form the final summary. 

\item \textsc{\textbf{BertSumExt}}~\cite{Liu2019TextSW}: A neural extractive model that fine-tunes \textsc{Bert}~\cite{Devlin2019BERTPO} language model on extractive summarization text. This model runs by appending \texttt{[CLS]} tokens to the start of each input sentence, and use the representations associated with \texttt{[CLS]} tokens to predict sentence importance. \texttt{[CLS]} is the classification head in \textsc{Bert} model that aggregates the contextualized embeddings of preceding tokens.

\item \textsc{\textbf{MatchSum}}~\cite{Zhong2020ExtractiveSA}: A state-of-the-art extractive summarization model that first composes candidate summaries given the salient set of source sentences scored by \textsc{BertSumExt} model, and then ranks them using the Siamese neural networks. Top-ranked candidate summary is retrieved as the final extractive summary of the post.

\item \textsc{\textbf{BertSumAbs}}~\cite{Liu2019TextSW}: The abstractive variant of \textsc{BertSum} framework, where the encoder is simply a \textsc{Bert} model, which is trained alongside a Transformers-based~\cite{Vaswani2017Att} decoder from scratch. 


\item \textsc{\textbf{BertSumExtAbs}} ~\cite{Liu2019TextSW}: A two-stage fine-tuned abstractive model that exploits a pre-trained \textsc{BertSumExt} summarizer (i.e., first stage) which is further fine-tuned along with a decoder on abstractive summarization task (i.e., second stage).

\item \textsc{\textbf{Bart}} ~\cite{Lewis2020BARTDS}: An abstractive model that is currently amongst the most powerful state-of-the-art summarization models. \bart{} extends the \textsc{Bert}'s intuition by adding up a couple of pre-training objectives including token deletion, text infilling, sentence permutation, and document rotation. Unlike \textsc{Bert}, \bart{} utilizes a pre-trained encoder-decoder framework for language generation task,  summarization being one of them.

\end{itemize}

\subsection{Implementation details}

We used Sumy package~\footnote{\url{https://github.com/miso-belica/sumy}} for running non-neural extractive models. To find extractive oracle labels (i.e., important sentences) of users' post, we ran a greedy labeling approach~\cite{Liu2019TextSW} over the entire set of source sentences and retrieved up to 3 sentences as the extractive summary. For \textsc{BertSum} models, we used the official codebase~\footnote{\url{https://github.com/nlpyang/PreSumm}} with {\fontfamily{cmtt}\selectfont BERT-base-uncased} and ran all of the models with default hyper-parameters as suggested by~\newcite{Liu2019TextSW} besides the learning rate of $1e-3$ and warmup steps of 5k. For \textsc{MatchSum}, we used  {\fontfamily{cmtt}\selectfont RoBERTa-base} as the encoder with the same default hyper-parameters as initialized in the original paper~\cite{Zhong2020ExtractiveSA} and candidate summaries of lengths 2 and 3.  We utilized Huggingface Transformers'~\cite{Wolf2020Transformers} implementation for training \bart{} model. All the experimented neural models were trained for 5 epochs and then the best checkpoint that attains the highest \rgl{} score during validation was picked for inference time. As the optimizer, we used AdamW~\cite{Loshchilov2019DecoupledWD} initialized with learning ratio of $3e-5$, $(\beta_1, \beta_2)= (0.9, 0.98)$, and a weight decay of 0.01. Cross-entropy loss was employed for all of the experimented models. We also utilized Weights \& Biases toolkit~\cite{wandb} to keep track of training and validation progress. 

\begin{table*}[t]
\renewcommand*{\arraystretch}{1.05}
\centering 
\begin{tabular}{lrrr}
\toprule
 Method                    & \textsc{RG-1}  & \textsc{RG-2}  & \textsc{RG-L}  \\
\midrule
 \textsc{Lead-2}             &  20.09    & 3.57  & 13.70     \\
 \textsc{OracleExt}             &  35.98    & 11.59  & 23.21 \\ \vspace{-0.3cm}
\\ \hdashline \vspace{-0.3cm}  \\
 \textsc{LSA}~\cite{Steinberger2004LSA}                  &  22.96    & 3.98  & 14.50     \\

 \textsc{LexRank}~\cite{Erkan2004LexRankGL}                        &  23.21 & 4.42  & 15.19 \\

 \textsc{BertsumExt}~\cite{Liu2019TextSW}                        & 24.64 & 5.83  &  16.66  \\
 
\textsc{MatchSum}~\cite{Zhong2020ExtractiveSA}                        & 26.29 & 6.32  &  17.12  \\ \vspace{-0.3cm} \\  \hdashline \vspace{-0.3cm}    \\ 

 \textsc{BertsumAbs}~\cite{Liu2019TextSW}              &  25.35 & 6.49  &  17.11 \\
 
  \textsc{BertsumExtAbs}~\cite{Liu2019TextSW}              & 25.75  &  6.80 & 17.48  \\
  
  \textsc{Bart}~\cite{Lewis2020BARTDS}              &  \textbf{29.13} & \textbf{7.98}  & \textbf{20.27}  \\


\bottomrule
\end{tabular}
\caption{\rgf{} results on \mentsum 's test set.}
\label{tab:final}
\end{table*}

\section{Results and discussion}
\label{sec:results}
Table \ref{tab:final} presents the performance of summarization models on the test set of \mentsum{} dataset in terms of widely adopted \textsc{Rouge (F1)} metric~\cite{Lin2004ROUGEAP} that measures the text overlap between the system-generated \tldr{} summaries and user-written \tldr{} summaries. As expected and was shown in the data analysis, \textsc{Lead-2} is unable to perform well on \mentsum{} dataset. As addressed earlier, this is due to scattered salient information throughout the post. 

Comparing extractive models with the abstractive ones, we notice that extractive models lag behind the best performing abstractive model (i.e., \bart) by a huge margin;
this is not surprising as summaries are paraphrased by users, 
as also confirmed by Figure \ref{fig:novel} in Section \ref{sec:dataset_analysis}.  Even when comparing abstractive variants of \textsc{BertSum} model (i.e., \textsc{BertSumAbs} and \textsc{BertSumExtAbs}) with the extractive models, we see a similar trend of gaining more significant improvements via abstractive models. 

While comparing the extractive models with each other, we observe that \textsc{BertSumExt} model considerably outperforms the other unsupervised models (i.e., \textsc{LSA} amd \textsc{LexRank}), though, it lags behind the \textsc{MatchSum} model as the best-perfomring extractive model. This shows the effect of learning salient sentences in supervised manner, accounting for the improvements of both \textsc{BertSumExt} and \textsc{MatchSum} models. Noting that \textsc{OracleExt}
shows the upper bound performance of extractive summarization models, it is noticeable that we see a large gap between the extractive systems and \textsc{OracleExt}'s performances. This observation is a clear justification that there is still a large room for improvement in extractive summarization setting, which can be further approached by future work.

While looking at the performances of abstractive summarization models, we see a remarkable improvement of \bart{} over other systems; particularly, with relative improvements of 13.12\% (RG-1), 17.35\% (RG-2), and 15.96\% (RG-L) compared to \textsc{BertSumExtAbs} model~\footnote{The relative improvement of system A over system B is calculated as $\frac{\text{Rouge}_{\text{A}} - \text{Rouge}_{\text{B}}}{\text{Rouge}_{\text{B}}}$}. This is expected as \bart{} uses pre-trained encoder-decoder framework, unlike abstractive models of \textsc{BertSum} that only employ the pre-trained encoder (i.e., \textsc{Bert}) while training the decoder from scratch. It is also observed that \textsc{BertSumExtAbs} model surpasses \textsc{BertSumAbs} model, showing the effectiveness of extractive objectives in abstractive summary generation process as also suggested by ~\newcite{Gehrmann2018BottomUpAS}.

\begin{figure}
    \centering
    \includegraphics[scale=0.35]{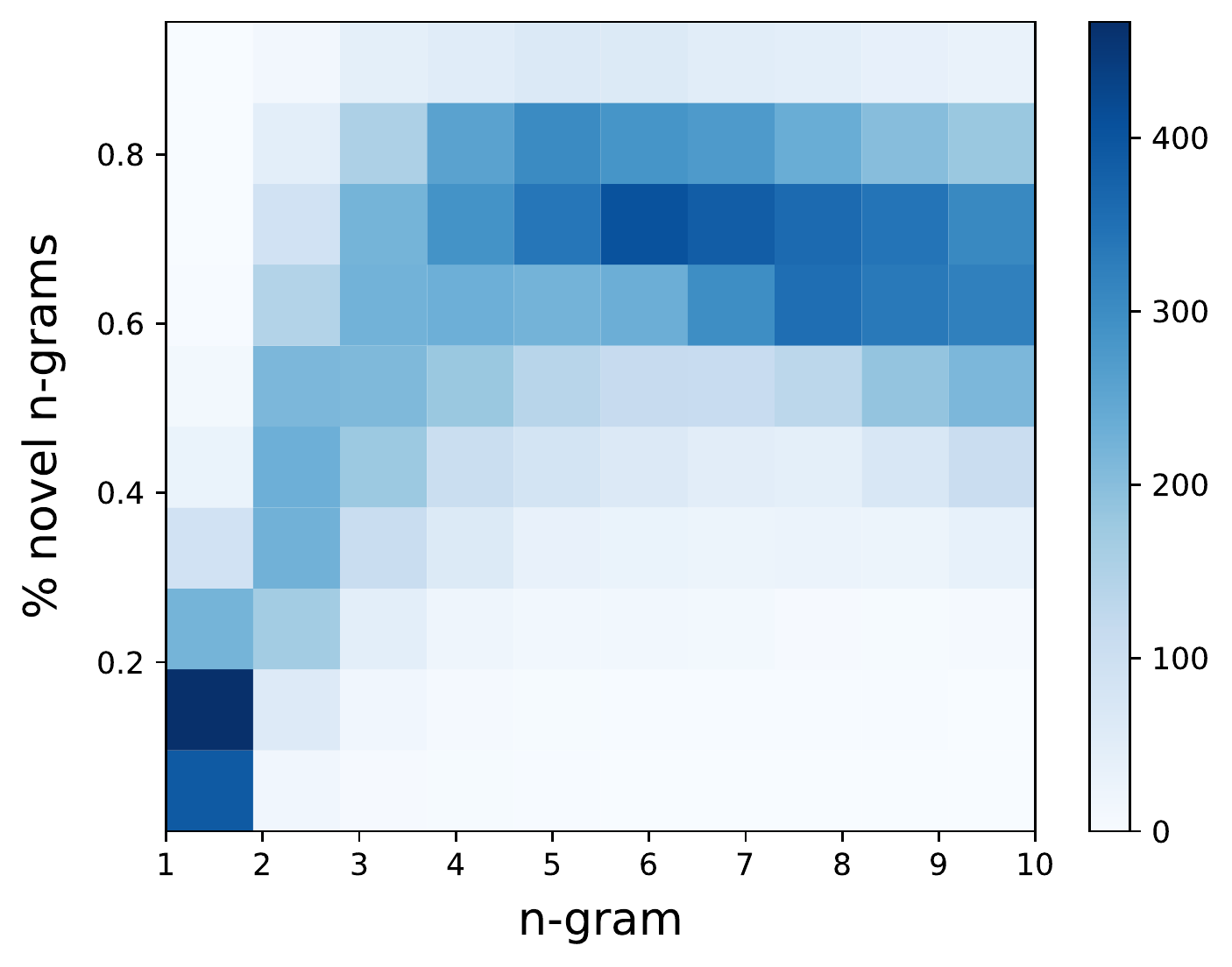}
    \caption{Percentage of novel n-grams across increasing word n-grams in \bart 's system-generated \tldr s on test set. The heat extent shows the frequency of word n-grams for a certain percentage bin.}
    \label{fig:bart-ngrams}
\end{figure}

Figure \ref{fig:bart-ngrams} demonstrates the percentage of novel n-grams in \bart -generated \tldr s over the test set.\footnote{The novel n-grams diagram on user-written \tldr s in test set is quite similar to Figure \ref{fig:novel} and is not shown due to space limitations.} As indicated, with increasing n-grams, the rate of novel words generation grows, showing that \bart{} does the abstraction by rephrasing/paraphrasing the source sentences. Also, compared to the user-written \tldr s, \bart{} does more extraction (on word level) than users do; however, it is still hard to attribute this observation to the quality of generated \tldr s as the goldness of ground-truth \tldr s may be influenced by the emotional state of the author at the time of writing as noted in Section \ref{sec:dataset_cons}.

\section{Human analysis}
\label{sec:human}
A few prior studies have recognized the limitations of widely used \textsc{Rouge} metric on qualitative evaluation as it is rather biased towards surface lexical similarities~\cite{Ng2015BetterSE,Cohan2016RevisitingSE}. To shed light on the qualities and limitations of the state-of-the-art summarizer (i.e., \bart), and provide insightful directions for future work, we carried out a human evaluation study. To this end, we randomly selected 100 posts along with their associated user-written and system-generated \tldr{} summaries from the test set. Following prior work~\cite{Grusky2018NewsroomAD,Zhang2020OptimizingTF,Cho2021StreamHoverLT,sotudeh-etal-2021-tldr9}, we defined three qualitative metrics: \textbf{(1) Fluency:} is the summary well-written and easy to understand?; \textbf{(2) Informativeness:} does the \tldr{} summary provide useful information (i.e., the most important information) about user's post?; \textbf{(3) Conciseness:} does the summary briefly provide comprehensive information (i.e., majority of important information) about user's post? Two human annotators evaluated the provided cases using 5-point Likert scale (1 = worst, 5 = best). For disagreement cases, where the assigned scores are 2 levels or more different from each other, a third annotator broke the disagreement and made the final decision. We then provided selected cases through the following multi-stage pipeline:

\begin{enumerate}
    \item In the first stage, we provided 100 posts with one randomly selected summary (from either user-written or system-generated), and asked the annotators to independently score the given summary in terms of the qualitative criteria.
    
    \item In the second stage, we provided the same 100 posts but with the other summary that was not shown to the annotators in the first stage, and asked the annotators to independently evaluate the given summary on the qualitative criteria. 
    
    \item In the final stage, we provided the same 100 posts with both summaries (i.e., user-written and system-generated) side-by-side, shuffled, non known to annotators and asked the annotators to specify which summary they prefer the most. The annotators were further asked to stipulate which major pieces of information are captured by or missing from each given summary. We will discuss the details of this stage in Section \ref{sec:error_anal} (i.e., Error Analysis).
\end{enumerate}

While stages (1) and (2) could have been combined into one stage, to avoid any bias in scoring these two summaries, we decided it was better that the annotators would score each of these two summaries independently without any comparison.  
It has to be mentioned that the order of summaries was shuffled; that is, the annotators were not aware of which of the two summaries they were evaluating. 

\begin{table}[t]
    \centering
    \scalebox{1.}{
    \begin{tabular}{lccc}
        \toprule
         System & Fluency & Info. & Conc.  \\
         \midrule
         \bart{} & \textbf{4.60} & \textbf{3.65} & \textbf{3.51}   \\
         User & {4.28} & 3.46  &  3.15 \\ 
         \bottomrule
    \end{tabular}
    }
    \caption{Results of the human evaluation comparing the systems in terms of {Fluency}, {Informativeness}, and {Conciseness}. Winning scores are shown in bold. Scores are in 5-point Likert scale (1=worst, 5=best).}
    \label{tab:human_evaluation}
\end{table}

\begin{table}[t]
    \centering
        \scalebox{1}{
    \begin{tabular}{lccc}
         \toprule
         System & Fluency & Info. & Conc.  \\
         \midrule
         \bart{} & 17.4\% & 28.2\% & 28.0\% \\
         User & 19.5\% & 25.2\% & 24.6\% \\
         \bottomrule
    \end{tabular}
    }
    
    \caption{System-wise Cohen's kappa inter-rater agreement.}

    \label{tab:agreement}
    \vspace{-0.1cm}
\end{table}

The evaluation scores are averaged and shown in Table \ref{tab:human_evaluation}. As shown, the \textsc{Bart} summarizer outperforms the user-written \tldr{} summaries across all metrics with relative improvements of 7.47\% (Fluency), 5.49\% (Informativeness), and 11.42\% (Conciseness). Observing the significant improvement gain by the state-of-the-art summarizer, it does seem interesting that contextualized language modelling based approaches such as \bart{} are becoming a firm touchstone to be compared with the human system. This finding is consistent with the observations that have been made recently by a prior work \cite{Fabbri2021SummEvalRS}. 

Table \ref{tab:agreement} reports the Cohen's kappa~\cite{Cohen1960ACO} inter-rater agreement for both system-generated and user-written \tldr{} summaries. With regard to the Cohen's kappa  range interpretation~\cite{McHugh2012InterraterRT}, the agreements obtained on informativeness and Conciseness metrics fall into the ``fair'' agreement, while there is a ``slight'' agreement observed on fluency. The slight agreement on fluency could be attributed to the subjective nature of this metric in evaluation process~\cite{Lee2021HumanEO}, leading to a high variability in assigned scores by the annotators. In addition to the overall promising performance of \bart, we observed that the nature of data is also impactful in the process of human study. In other words, users in social media are free to publish their content of discussion in whatever style they like to as there is usually not any supervision over the posted content in terms of quality, which leads to having gibberish \tldr s in some cases. 

As the results of stage (3) of the evaluation, the annotators reported that they  prefer system-generated summaries in 59\% of the cases, and user-written summaries in 41\% of the cases with an agreement rate of 59.6\% (moderate).

\section{Error analysis}
\label{sec:error_anal}

In this section, we showcase where the system summarizer lacks in or improves compared to the user-written summaries in more detail.  \textcolor{black}{As reported by the annotators, the distribution of qualitative criteria for determining which summary is more preferable is shown in Table \ref{tab:criteria_selection}. For instance, annotators pick the \bart-generated summaries in 59\% of cases. In 64\% of \bart 's \textit{win} cases, the system-generated summary was preferred due to its informativeness as compared to the user-written summary. As indicated, out of 3 evaluation metrics, informativeness is the most dominant one in the selection of \bart -generated and user-written summaries.} 

\begin{table}[t]
    \centering
    \begin{tabular}{lrrrr}
    \toprule
         System & Win rate & Fluency & Info. & Conc.  \\
         \midrule
         \bart{}  & 59\% & 17\%& 64\%& 19\%\\
         User & 41\% & 8\%& 81\%& 11\%\\
    \bottomrule
    \end{tabular}
    \caption{Distribution of leading criteria for summary selection in human evaluation process.}
    \label{tab:criteria_selection}
\end{table}

    

We report the findings for each of the three metrics of fluency, informativeness, and conciseness~\footnote{Examples are truncated in the paper for extra privacy of the users.}: 

\begin{itemize}[leftmargin=*,label={-}]

    \item \textbf{Fluency. } Few reasons contributed to annotators' lower score for user-written \tldr s in terms of fluency. One of them is grammatical issues in the users' language in user-written \tldr{} summaries. 
    Examples of such grammatical error is 
    ``{\fontfamily{cmtt}\selectfont ...good friend and ex girlfriend \textbf{has}...}''. Another observation showed the complex structure of the user-written sentence which makes the summary less understandable such as in ``{\fontfamily{cmtt}\selectfont ...she can't be arsed trying to communicate with myself...}''. 
    Interestingly, the system-generated \tldr s did not suffer from grammatical errors. This might be due to the fact that the system summarizer is exposed to the structured, grammatical, and fluent textual data from books and Wikipedia during pre-training, and hence with further fine-tuning on the \mentsum{} dataset, it is yet able to produce understandable \tldr s. 
    On the other hand, for the cases where system-generated summaries underperform the user-written ones, we observed examples of repetition such as in ``{\fontfamily{cmtt}\selectfont i'm having massive mood swings and mood swings...}'', and complexity without specifying sentence boundaries such as in ``{\fontfamily{cmtt}\selectfont ...i was @ageX i am @ageY now and i haven't been able to find a therapist until my @ageZ birthday and i want to offer myself...}''.
    
    \item \textbf{Informativeness. } We observed that the users provide some history of their mental health and disorders within their posts. It is important for these vital information to be captured in the system-generated summaries. 
    Interestingly, the system summarizer attends more to such mental disorders and includes them into the generated summaries.  
    For instance, in the system-generated summary  ``{\fontfamily{cmtt}\selectfont ...diagnosed with anxiety, GAD, and agoraphobia...}'' 
    all mental illnesses of the user are captured.
    However, the user-written \tldr{} ``{\fontfamily{cmtt}\selectfont ...was diagnosed with anxiety...}'' does not include ``{\fontfamily{cmtt}\selectfont GAD, and agoraphobia}''. 
    Furthermore, we observed that parts of user-written summaries contain information that is not directly mentioned in the user's post, but was inferred from it. Such cases are still a challenge for the system summarizers. 
    As an example, consider user-written \tldr{} of ``{\fontfamily{cmtt}\selectfont ...I'm holding it all in which is making me and everyone around me worse...}'', which conveys information that is not directly mentioned in the post, but could be inferred from the post. While user-written \tldr{} contains such salient information, the system-generated \tldr{} was not able to infer and generate that. In underperformed cases, we observed that the model focuses on a few (one or two) important points of the user’s post, while the user-written \tldr{} includes more important points from diverse parts of the user’s post.
    
    \item \textbf{Conciseness. } \textcolor{black}{The aforementioned limitations of informativeness affect the conciseness negatively. That is, some of important information that are required to be in a concise summary are missed.} In particular, for the cases in which user-written \tldr s are scored lower than 
    system-generated ones in terms of conciseness, the user-written \tldr s are verbose and very detailed.
    However, the system is able to skip unnecessary information, focus on discerning important information, and verbalize them within a comparably shorter text. 
    For instance, consider system-generated \tldr{} ``{\fontfamily{cmtt}\selectfont ...been on meds and seeing a psych for a while, now finally going to a dr for help...}'' in comparison with its user-written \tldr{} ``{\fontfamily{cmtt}\selectfont ...saw a psych about it but then stopped. felt like garbage last year. seeing a doctor about it now. happy hopeful noises...}''. As seen, the system  verbalizes the important points using a comparably briefer usage of language.

\end{itemize}

\section{Summary and conclusion}

Mental health and illnesses have become a major challenge in public health. The prevalence of user-generated and user-curated content in online social media platforms has provided a large amount of accessible information for those who seek to study user behavioural patterns in social settings. While most previous research studies in mental health domain  have focused on developing datasets for the classification tasks to triage and detect the type and severity of mental health concerns, there has not been any dataset to support the research in summarizing mental health related social media posts.  
In this paper, we introduced our large-scale novel dataset resource, \mentsum{} for the task of summarization of users' mental health posts on Reddit social media. Although the dataset is gathered from 43 publicly available Reddit subreddits, it will be provided to researchers through a Data Usage Agreement (DUA) to protect the privacy of the users. 
Our detailed dataset analyses
 revealed characteristics such as weak lead bias, strong abstractiveness, and extractiveness. We further evaluated various state-of-the-art extractive and abstractive summarization models on \mentsum{} dataset. This study showed that 
 while the abstractive method (i.e., \bart) outperformed the extractive models, 
the gap between the extractive models' performance and \textsc{OracleExt} (i.e., the upper bound performance of extractive methods) indicates a large room for improvement in extractive settings, calling for future research. 
Our 
human evaluation 
over 100 randomly selected post summary pairs in terms of fluency, informativeness, and conciseness of system-generated and user-written summaries revealed interesting information; 
the system-generated summaries were preferred on average over 
user-written ones,  
showing the promising performance of contextualized language modelling based summarization approaches. 
Finally, we provided an error analysis to show the current challenges. We hope our provided \mentsum{} dataset paves the path for further exploration in summarization research of social media mental health posts.  

\section{Bibliographical References}\label{reference}

\bibliographystyle{lrec2022-bib}
\bibliography{lrec2022-example}
\bibliographystylelanguageresource{lrec2022-bib}
\bibliographylanguageresource{languageresource}

\end{document}